\documentclass{article}
\usepackage{spconf,amsmath,graphicx,amssymb,cite,booktabs,url}
\usepackage{multirow}
 \usepackage{enumitem}
\setlist[itemize]{noitemsep} 
\newcommand{\eegtrans}{EEGXF}

\title{TEMPORAL CONTEXT AND ARCHITECTURE:\\A BENCHMARK FOR NATURALISTIC EEG DECODING}

\name{Mehmet Ergezer}
\address{School of Computing \& Data Science, Wentworth Institute of Technology \\
Boston, MA USA \\
         Email: ergezerm@wit.edu}

\begin{document}

\maketitle

\begin{abstract}
We study how model architecture and temporal context interact in naturalistic EEG decoding. Using the HBN movie‑watching dataset, we benchmark five architectures, CNN, LSTM, a stabilized Transformer (\eegtrans{}), S4, and S5, on a 4-class task across segment lengths from 8 s to 128 s. 
% we benchmark CNN, LSTM, a stabilized Transformer (\eegtrans{}), and a structured state‑space model (S5) on a 4‑class task across segment lengths from 8s to 128s. 
Accuracy improves with longer context: at 64 s, S5 reaches 98.7\%$\pm$0.6 and CNN 98.3\%$\pm$0.3, while S5 uses $\sim$20$\times$ fewer parameters than CNN. To probe real‑world robustness, we evaluate zero‑shot cross‑frequency shifts, cross‑task OOD inputs, and leave‑one‑subject‑out generalization. S5 achieves stronger cross‑subject accuracy but makes over‑confident errors on OOD tasks; \eegtrans{} is more conservative and stable under frequency shifts, though less calibrated in‑distribution. These results reveal a practical efficiency–robustness trade‑off: S5 for parameter‑efficient peak accuracy; \eegtrans{} when robustness and conservative uncertainty are critical.

\end{abstract}


\section{Introduction}
\label{sec:intro}

Decoding brain activity from EEG during naturalistic tasks like movie-watching is a challenging problem, requiring models that capture long-range temporal dependencies from noisy, continuous recordings~\cite{shirazi2024hbn}. While deep learning has been widely applied to EEG, traditional architectures face fundamental trade-offs. CNNs are efficient but constrained by local receptive fields~\cite{roy2019deep, schirrmeister2017deep}, while Transformers offer global context via attention~\cite{vaswani2017attention} but suffer from quadratic scaling in time and memory.

To address these limitations, Structured State-Space Models (SSMs) like S4~\cite{gu2022efficientlymodelinglongsequences} and S5/Mamba~\cite{gu2023mamba} have emerged, offering a compelling balance of long-range modeling and near-linear computational complexity. Recent studies have applied SSMs to EEG foundation models~\cite{wang2025eegmamba}, drowsiness detection~\cite{siddhad2024drowzee}, and clinical tasks~\cite{tran2024eegssm, li2025dgmamba}. However, a systematic comparison of how these modern architectures perform against varying temporal context lengths has been missing.

In this work, we investigate the interplay between model architecture and temporal context. We hypothesize that sequence models like S5 and Transformers benefit more from longer EEG segments than local models like CNNs. To test this, we benchmark five architectures on a 4-class naturalistic decoding task using HBN data across segments from 8\,s to 128\,s. While hybrid architectures like EEG-Conformer~\cite{song2023eeg} achieve state-of-the-art performance by layering mechanisms, we focus on \textit{fundamental} architectural classes (CNN, RNN, SSM, Transformer) to isolate the impact of specific inductive biases. To move beyond simple accuracy, we introduce a suite of challenging generalization tests across unseen tasks, frequencies, and subjects. This evaluation characterizes a fundamental trade-off: S5 offers high speed and accuracy but is prone to overconfidence on out-of-distribution data, whereas \eegtrans{} provides greater robustness.

% To test this, we benchmark five architectures on a 4-class naturalistic decoding task using HBN data across segments from 8s to 256s. To ensure a robust comparison, we also develop \eegtrans{}, a modernized Transformer baseline for EEG. Our contributions are: (1) a systematic evaluation of five architectures against temporal context length; (2) the introduction of \eegtrans{} as a strong baseline; (3) novel cross-frequency and leave-one-subject-out (LOSO) generalization tests; and (4) the characterization of a critical performance-robustness trade-off between S5 and \eegtrans.

% \paragraph*{Our Contributions}
% \begin{itemize}
%     \item We present the first systematic evaluation of five neural architectures across EEG segment lengths from 8s to 256s.
%     \item We introduce \eegtrans{}, a modernized and robust Transformer baseline for EEG decoding.
%     \item We evaluate generalization using three robust tests: (1) zero-shot cross-task inference on unseen cognitive states, (2) cross-frequency downsampling, and (3) leave-one-subject-out (LOSO) subject transfer.
%     \item We identify and characterize a fundamental trade-off: S5 offers state-of-the-art speed and accuracy but is prone to overconfident errors on out-of-distribution data, whereas \eegtrans{} provides greater robustness and safer uncertainty handling at a higher computational cost.
% \end{itemize}

% BibTeX entries to be added:
% - EEGMamba
% - DrowzEE-G-Mamba
% - EEG-SSM (Tran et al. 2024)
% - DG-Mamba (Li et al. 2025)

\section{Dataset and Preprocessing}
\label{sec:data}

We utilize EEG data from the Healthy Brain Network (HBN) initiative \cite{shirazi2024hbn}, specifically recordings from the movie-watching tasks (``Despicable Me,'' ``Diary of a Wimpy Kid,'' ``The Present'') and a resting-state baseline.
The raw EEG data was downsampled to 250\,Hz, and the first 64 channels were selected. Our preprocessing pipeline consisted of:
\begin{enumerate}[itemsep=0cm,parsep=0.1cm,leftmargin=*]
    \item \textbf{Filtering:} Signals were band-passed (1-50Hz). For downsampling experiments, an anti-aliasing low-pass filter (cutoff $0.45 \times$ target rate) was applied before resampling.
    \item \textbf{Re-referencing:} Common average reference.
    \item \textbf{Split \& Segmentation:} To prevent data leakage, continuous recordings were \textbf{first split} into training (60\%), validation (20\%), and test (20\%) sets. Only then were signals segmented into $T$-second windows with 50\% overlap. This ensures no overlapping windows span across train-test splits.
    \item \textbf{Normalization:} Each segment was standardized on a per-channel basis (zero mean, unit variance).
\end{enumerate}

Our benchmark uses data from 40 subjects, yielding over 15,000 EEG segments for the 4-class classification task.

\section{Model Architectures}
\label{sec:models}

We benchmark five distinct architectures to evaluate the impact of inductive bias: a temporal CNN, a bidirectional LSTM, a vanilla Transformer, our stabilized \eegtrans{}, and the structured state-space model S5. All models accept input tensors of shape (batch, 64, $T$), where $T$ varies with segment length (e.g., $T=2000$ for 8\,s).

\subsection{Structured State-Space Model (S5)}
Our S5 classifier builds on recent advances in state-space modeling. It is governed by the continuous-time linear system:
\begin{align}
\mathbf{h}'(t) &= \mathbf{A} \mathbf{h}(t) + \mathbf{B} \mathbf{u}(t) \\
\mathbf{y}(t) &= \mathbf{C} \mathbf{h}(t) + \mathbf{D} \mathbf{u}(t)
\end{align}
In contrast to S4, which uses FFT-based convolutions, S5 directly diagonalizes the state matrix $\mathbf{A}$ to enable an efficient parallel scan.
Our implementation uses a stack of 3 bidirectional S5 blocks (hidden dim=192), global average pooling, and a linear classifier. This configuration (183K parameters) allows us to test the efficiency of pure sequence modeling without the quadratic cost of attention.

\subsection{\eegtrans{}: A Stabilized Transformer Baseline} \label{sec:\eegtrans}
Standard Transformers often suffer from training instability on noisy EEG data. To create a robust baseline, we developed \eegtrans{}, a stabilized EEG transformer. Key modifications include a simplified input projection with ReLU activation, multiple BatchNorm layers for stabilization, a conservative encoder with \textit{norm\_first=True}, and high-variance weight initialization. Based on hyperparameter sweeps, we utilized a compact configuration (2 layers, 4 heads, $d=128$, dropout=0.1) which outperformed larger variants. To aggregate temporal features, \eegtrans{} replaces global pooling with a learned attention pooling mechanism.

The CNN (3 layers, 1D convolution) and LSTM (2 layers, bidirectional) serve as standard local and recurrent baselines, respectively. We release all code for reproducibility.\footnote{Code available at: \url{https://github.com/memoatwit/eeg-s5-benchmark/}}

\section{Experiments}
\label{sec:exp}

Models were trained for up to 100 epochs using the AdamW optimizer, a `ReduceLROnPlateau` scheduler, and early stopping. We used gradient clipping and EMA to stabilize training. %Results for the 64s segment are reported as mean~$\pm$~std over three random seeds; due to  computational cost, other results are from a single seed. 
Results for 32 s, 64 s, and 128 s are reported as mean~$\pm$~std over 3 seeds (CNN/\eegtrans{}/S5); 8s and 16s are single‑seed due to cost.
Reported training times are wall‑clock to early‑stopped convergence on a single NVIDIA RTX 3090 Ti. %, averaged over 3 seeds for 64s. 
Unless otherwise noted, Figures~\ref{fig:segcurve}-\ref{fig:efficiency_scatter} show single-seed results for five architectures (CNN, \eegtrans{}, LSTM, S4, S5), whereas Table~\ref{tab:main} reports mean $\pm$ std over three seeds for our focal models (CNN, \eegtrans{}, S5) at 32/64/128 s to conserve space.

\section{Results}
\label{sec:results}
\begin{figure}[t]
  \centering
    \includegraphics[width=0.95\linewidth]{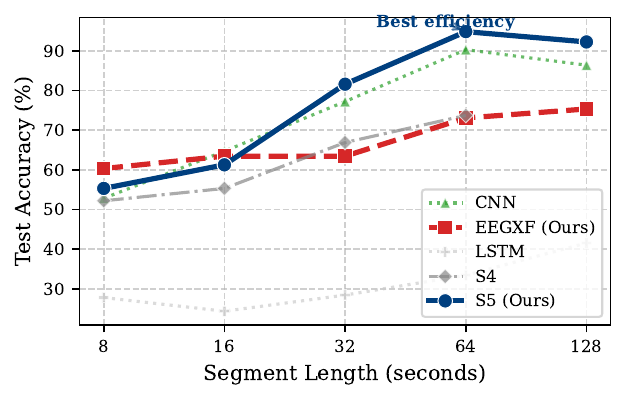}
  % \caption{Test accuracy versus segment length. S5 and CNN demonstrate the strongest performance scaling with temporal context. The x-axis is log-scaled.}
\caption{Test accuracy versus segment length (8–128 s). Single-seed comparison of five architectures (CNN, \eegtrans{}, LSTM, S4, S5).  S5 and CNN demonstrate the strongest performance scaling with temporal context; x-axis log-scaled.}
  %Test accuracy versus input segment length for EEG classification models. Our proposed S5 and \eegtrans{} architectures demonstrate superior performance across all temporal contexts, with S5 achieving optimal efficiency at longer segment lengths. The x-axis uses logarithmic scaling.}
  \label{fig:segcurve}
\end{figure}

Our primary results on subject-mixed splits are summarized in Figure~\ref{fig:segcurve} and Table~\ref{tab:main}\footnote{Subject-mixed: train/test split includes overlapping subjects. Subject-independent results are detailed in Sections 5.2, 5.3.}. 
Accuracy improves with longer context: at 64 s, S5 reaches $98.7\%\pm0.6$ and CNN $98.3\%\pm0.3$, 
while S5 uses $\sim$20$\times$ fewer parameters than CNN. 
CNN remains the fastest wall-clock at 64 s (9.8 $\pm$ 0.03 min), and S5 and \eegtrans{} have comparable training times (25.8 $\pm$ 6.2 vs.\ 25.1 $\pm$ 0.1 min). 

% Our primary results on subject-mixed splits are summarized in Figure 1 and Table 1. The experiment confirms our hypothesis: models with long-range capacity benefit more from temporal context. The CNN baseline is a remarkably strong performer, achieving a peak accuracy of 97.2\%~$\pm$~0.9 at 64s. S5 scales nearly as well, peaking at 95.2\%~$\pm$~1.7. %, but does so with over 20x fewer parameters and significantly faster training. Our engineered \eegtrans{} also scales effectively but peaks lower than the other two main contenders.
% S5 trains slightly faster than \eegtrans{}  and is far more parameter‑efficient than CNN; CNN remains the fastest wall‑clock at 64s.

\begin{figure}[t]
  \centering
  \includegraphics[width=0.98\linewidth]{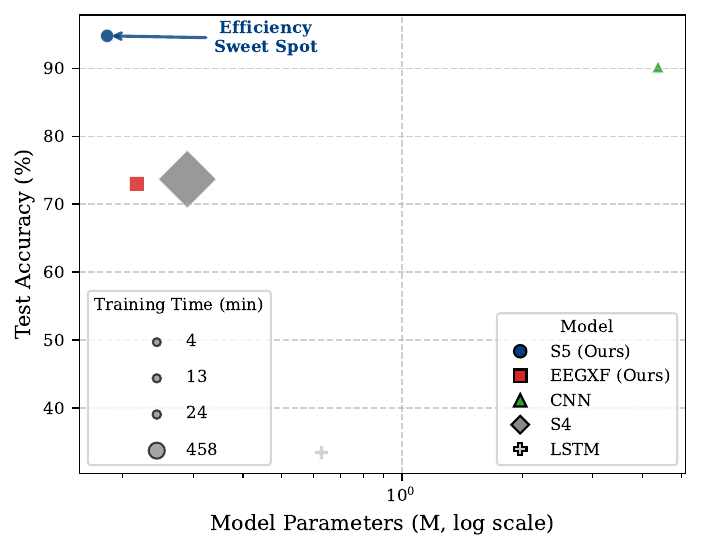}
  % \caption{ Performance–efficiency trade‑off at 64s. Accuracy vs parameters (log‑scale); marker size $\underset{\sim}{\propto}$ mean training time (min). CNN is fastest, while S5 attains state‑of‑the‑art accuracy with \textbf{$\sim20\times$} fewer parameters than CNN; EEGXF is comparable in training time to S5 but less accurate.}
  \caption{Performance–efficiency trade-off at 64 s. Single-seed comparison of five architectures; accuracy vs.\ parameters (log-scale); marker size $\propto$ training time (min). CNN is fastest (4.4 M parameters) while S5 (0.2 M parameters) attains state-of-the-art accuracy with $\sim$20$\times$ fewer parameters than CNN; \eegtrans{} is comparable in training time but less accurate.}
  %\caption{Performance-efficiency trade-off analysis at 64-second segments. Test accuracy is plotted against model parameters (millions, log scale) with marker size proportional to training time. S5 achieves the optimal balance of high accuracy ($>90\%$) with minimal parameters ($<0.2M$), demonstrating superior efficiency compared to the larger CNN and slower \eegtrans{}}
  \label{fig:efficiency_scatter}
\end{figure}

The traditional LSTM and our simplified S4 baselines were not competitive. The LSTM failed to converge, and the S4 model, while reaching 73.7\% accuracy at 64 s, was prohibitively slow; details in Fig. 2.
%over 7 hours
Figure~\ref{fig:efficiency_scatter} further illustrates the accuracy-efficiency trade-offs, showing S5 and CNN as the most efficient high-performing architectures.

\begin{table}[t]
\centering
\caption{Performance on 4-class subject-mixed EEG classification.
32/64/128 s are mean$\pm$std over 3 seeds; 8/16s are single-seed.}
% Results for 32/64/128s are averaged over 3 seeds; 8s and 16s are single-seed.
\label{tab:main}
\resizebox{\columnwidth}{!}{%
\begin{tabular}{lcccc}
\toprule
Model & Seg (s) & Acc. (\%) & F1 & Time (min) \\
\midrule
\multirow{5}{*}{\shortstack{S5 \\ (SSM)}}
& 8   & 55.3 & 0.55 & 3.9 \\
& 16  & 61.3 & 0.61 & 5.1 \\
& 32  & $\mathbf{94.4}\pm0.7$ & $0.93\pm0.013$ & $19.9\pm2.2$ \\
& 64  & $\mathbf{98.7}\pm0.6$ & $0.98\pm0.011$ & $25.8\pm6.2$ \\
& 128 & $\mathbf{95.8}\pm0.6$ & $0.93\pm0.007$ & $17.8\pm4.8$ \\
\midrule
\multirow{5}{*}{\shortstack{EEGXF \\ (Transformer)}}
& 8   & 60.3 & 0.61 & 16.2 \\
& 16  & 63.4 & 0.64 & 22.9 \\
& 32  & $80.1\pm2.0$ & $0.75\pm0.029$ & $35.8\pm0.3$ \\
& 64  & $81.8\pm1.8$ & $0.73\pm0.019$ & $25.1\pm0.1$ \\
& 128 & $76.7\pm1.5$ & $0.53\pm0.032$ & $6.3\pm2.6$ \\
\midrule
\multirow{5}{*}{\shortstack{CNN \\ (Local)}}
& 8   & 52.8 & 0.53 & 0.9 \\
& 16  & 64.7 & 0.65 & 1.2 \\
& 32  & $94.5\pm0.3$ & $0.93\pm0.004$ & $7.5\pm0.02$ \\
& 64  & $98.3\pm0.3$ & $0.98\pm0.004$ & $9.8\pm0.03$ \\
& 128 & $89.2\pm1.5$ & $0.68\pm0.097$ & $6.4\pm0.06$ \\
\bottomrule
\end{tabular}
}
\end{table}

% \begin{table}[t]
%   \centering
%   \caption{Performance on 4-class subject-mixed EEG classification. S5 and CNN excel at longer contexts, while S5 remains the most parameter-efficient.}
%   \label{tab:results}
%   \resizebox{\columnwidth}{!}{%
%   \begin{tabular}{lrrrrr}
%     \toprule
%     \textbf{Model} & \textbf{Seg (s)} & \textbf{Acc. (\%)} & \textbf{F1} & \textbf{Time (min)} & \textbf{Params} \\
%     \midrule
%     \textbf{S5} & 8 & 55.3 & 0.553 & 3.9 & \textbf{183K} \\
%     (SSM) & 16 & 61.3 & 0.614 & 5.1 & \textbf{183K} \\
%      & 32 & 81.6 & 0.814 & 8.0 & \textbf{183K} \\
%      & 64 & \textbf{94.8} & \textbf{0.949} & 12.8 & \textbf{183K} \\
%      & 128 & \textbf{92.2} & \textbf{0.895} & 9.6 & \textbf{183K} \\
%     \midrule
%     EEGXF & 8 & 60.3 & 0.605 & 16.2 & 217K \\
%     (Transformer) & 16 & 63.4 & 0.635 & 22.9 & 217K \\
%      & 32 & 63.4 & 0.632 & 13.4 & 217K \\
%      & 64 & 73.1 & 0.731 & 24.1 & 217K \\
%      & 128 & 75.3 & 0.593 & 9.9 & 217K \\
%     \midrule
%     CNN & 8 & 52.8 & 0.531 & \textbf{0.9} & 4.4M \\
%     (Local Baseline) & 16 & 64.7 & 0.650 & \textbf{1.2} & 4.4M \\
%      & 32 & 77.2 & 0.769 & \textbf{2.5} & 4.4M \\
%      & 64 & 90.3 & 0.900 & \textbf{4.0} & 4.4M \\
%      & 128 & 86.4 & 0.639 & \textbf{3.8} & 4.4M \\
%     \bottomrule
%   \end{tabular}
%   }
% \end{table}

% ... after Table 1 and its footnotetext ...

To further analyze classification behavior, we examined the confusion patterns between the three movie classes at the 64 s segment length (Table~\ref{tab:confusion_summary}). Both S5 and CNN demonstrate high precision, misclassifying only 2.7\% of movie segments. In contrast, \eegtrans{} exhibits significant confusion between movies (32.4\%), explaining its lower F1-score and highlighting its difficulty in separating the nuanced features of the different naturalistic stimuli compared to S5 and CNN.

\begin{table}[h]
\centering
\caption{Inter-movie confusion analysis at 64 s segments. The metric shows the percentage of movie segments incorrectly classified as a different movie.}
\label{tab:confusion_summary}
% \resizebox{0.85\columnwidth}{!}{%
\begin{tabular}{lc}
\toprule
\textbf{Model} & \textbf{Movie Confusion Rate (\%) $\downarrow$} \\
\midrule
S5 & \textbf{2.7\%} \\
CNN & \textbf{2.7\%} \\
\eegtrans{} & 32.4\% \\
\bottomrule
\end{tabular}
% }
\end{table}

% ... Continue with the rest of the Results section ...

Figure~\ref{fig:efficiency_scatter} displays the performance vs efficiency across architectures. 
S5 models occupy the top-left region of the efficiency-accuracy space, showing they achieve strong performance with minimal resources. In contrast, \eegtrans{} models trade robustness for compute cost.

% \subsection{Subject-Mixed vs. Subject-Independent Evaluation}
% A critical consideration in EEG decoding is the data splitting strategy. Our main results (Table~\ref{tab:results}) use a \textbf{subject-mixed} split, where segments from all subjects are pooled before being divided into training and test sets. While common, this may inflate performance by allowing models to learn subject-specific neural patterns.

% To provide a more realistic assessment of generalization, we also employ \textbf{subject-independent} splits in our cross-frequency and leave-one-subject-out (LOSO) evaluations. In these setups, the test set contains only subjects unseen during training. As we will show, this leads to a significant drop in accuracy (e.g., S5 performance falls from its subject-mixed peak of 94.8\% (at 64s) to 53.1\% in subject-independent tests (at 32s), highlighting the immense challenge of cross-subject generalization and the importance of reporting results under both conditions.

\subsection{Cross-Frequency Robustness}
\begin{figure}[t]
  \centering
  \includegraphics[trim={0 0 0 1.8cm},clip, width=0.98\linewidth]{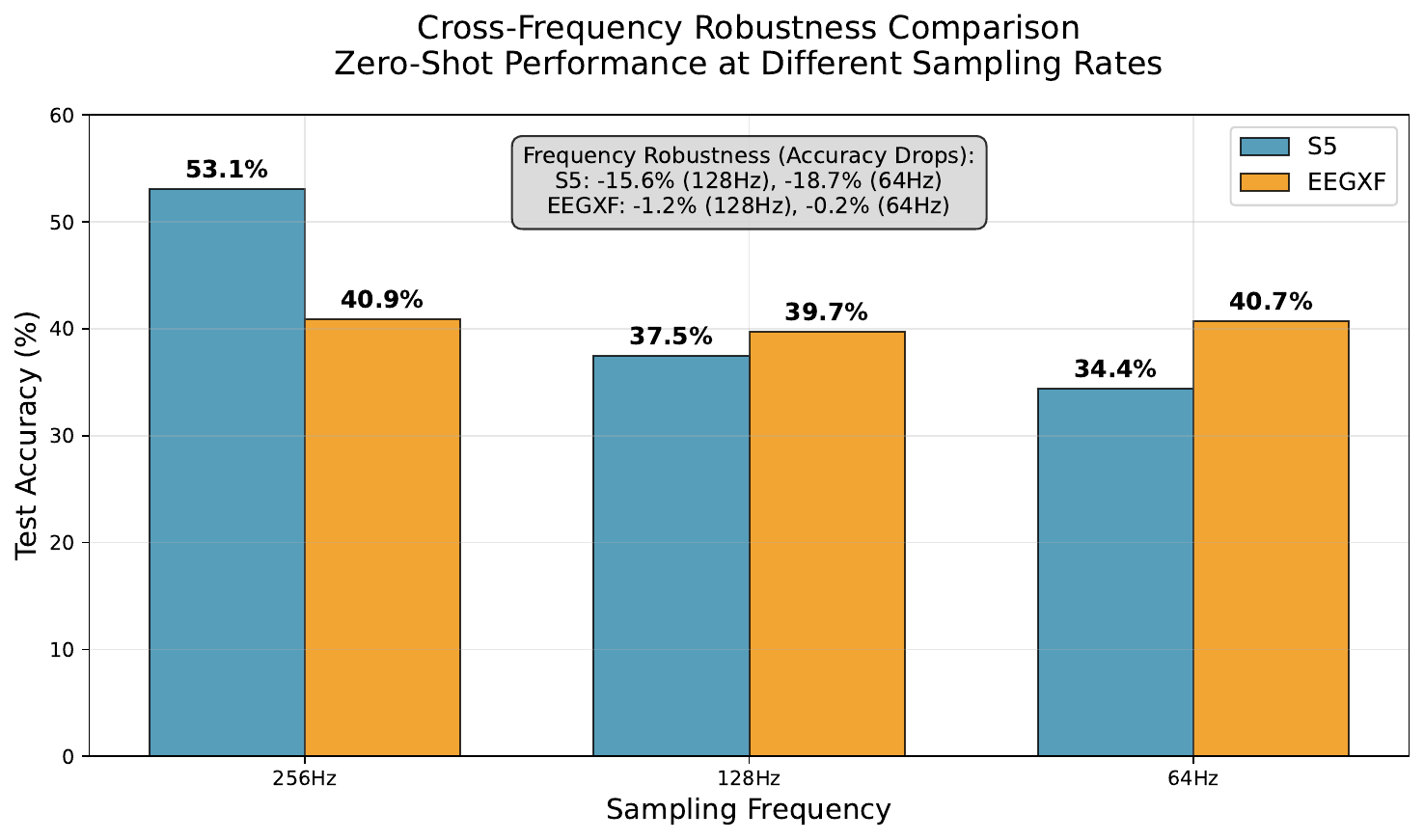}
  \caption{Zero-shot generalization performance at different sampling frequencies. 
  S5 shows a large drop in accuracy when tested on downsampled EEG, while \eegtrans{} 
  remains robust. This highlights a speed-accuracy-robustness trade-off across architectures.}
  \label{fig:crossfreq_bar}
\end{figure}

To assess generalization across EEG sampling rates, we conducted a zero-shot evaluation on subject-independent splits. Models were trained at 250Hz and tested on signals downsampled to 128Hz and 64Hz using FIR filtering. The results are shown in Table~\ref{tab:crossfreq} and Figure~\ref{fig:crossfreq_bar}. S5's accuracy drops significantly (over 18 percentage points), highlighting a sensitivity to temporal resolution. In contrast, \eegtrans{} remains remarkably stable, with an accuracy drop of less than 2\%. This reveals a key trade-off: S5 offers superior peak performance and speed, but \eegtrans{} is preferable for applications with variable sampling rates.

% This is Table 2
\begin{table}[h]
\centering
\caption{Zero-shot accuracy at different sampling rates.}
\label{tab:crossfreq}
% \resizebox{0.8\columnwidth}{!}{%
\begin{tabular}{lccc}
\toprule
\textbf{Model} & \textbf{250Hz} & \textbf{128Hz} & \textbf{64Hz} \\
\midrule
S5 & 53.1\% & 37.5\% & 34.4\% \\
\eegtrans & 40.9\% & 39.7\% & 40.7\% \\
\bottomrule
\end{tabular}
% }
\end{table}

Figure~\ref{fig:crossfreq_bar} highlights the trade-off between speed and robustness: S5 trains faster and achieves higher peak accuracy, but \eegtrans{} is more stable under sampling frequency changes.

\subsection{Leave-One-Subject-Out Generalization}
To rigorously assess subject-independent generalization, we conducted a full leave-one-subject-out (LOSO) evaluation on 32 s segments. We ran 60 LOSO folds (multiple sessions per subject); significance tests used the 20 folds available for both models. The results (Table~\ref{tab:loso_results}) confirm \textbf{S5's architecture is superior for cross-subject generalization}, achieving a mean accuracy of 55.9\%. While these scores are substantially lower than subject-mixed results and exhibit high variance (accuracies per fold ranged from 21\% to 100\%), they highlight the profound difficulty of cross-subject EEG decoding. 

\begin{table}[h!]
\centering
\caption{Leave-One-Subject-Out (LOSO) performance on 32 s segments across 40 subjects. S5 demonstrates stronger cross-subject generalization.}
% \caption{Leave-One-Subject-Out (LOSO) performance on 32s segments. S5 demonstrates significantly stronger cross-subject generalization than the Transformer baseline.}
\label{tab:loso_results}
\resizebox{\linewidth}{!}{%
\begin{tabular}{lcccc}
\toprule
\textbf{Model} & \textbf{Mean Acc. (\%)} & \textbf{Std. Dev.} & \textbf{Folds Run} & \textbf{Mean Time/Fold} \\
\midrule
\textbf{S5} & \textbf{55.9} & 15.9 & 60 / 60 & \textasciitilde41 min \\
\eegtrans{} & 48.4 & 12.1 & 20 / 60 & \textasciitilde52 min \\
\bottomrule
\end{tabular}
}
\end{table}

\textbf{Statistical significance.}
Across the 20 matched LOSO folds, S5 outperformed \eegtrans{} by $10.6 \pm 15.1$\,pp in accuracy ($t_{19} = 3.13$, $p = 0.0055$) and $0.25 \pm 0.16$ in F1-score ($t_{19} = 7.05$, $p = 1.0 \times 10^{-6}$), with medium-to-large effect sizes.
A McNemar test confirmed that S5 was significantly more likely to exceed 50\% accuracy across folds ($\chi^2 = 12.0$, $p = 4.9 \times 10^{-4}$).
These findings reinforce that S5 generalizes better than \eegtrans{} despite its overconfidence issues.

\subsection{Zero-Shot Cross-Task Generalization}
To test how models behave on unseen cognitive states, we evaluated our movie-trained models on three out-of-distribution (OOD) active tasks ($N=245$). The results reveal a striking difference in generalization strategy, supported by calibration metrics on the in-distribution test set (Table~\ref{tab:calibration}).

\textbf{S5 exhibited overconfident specialization.} While it is exceptionally well-calibrated on in-distribution data, achieving the best NLL and Brier and near‑best ECE (on par with CNN), it consistently misclassified all OOD tasks as `Movie 3' with a high mean confidence of 60.0\% (Table~\ref{tab:cross_task}). Its failure to express uncertainty on novel inputs is a critical limitation.

In stark contrast, \textbf{\eegtrans{} demonstrated a ``conservative collapse.''} Although less calibrated on in-distribution data, it correctly identified the OOD tasks as unfamiliar by defaulting to its most neutral known class (`Resting State') with an appropriately low confidence of 26.0\%. This behavior is not a failure but a sign of robust uncertainty quantification, a vital property for safe real-world deployment.

% \textbf{S5 exhibited overconfident specialization}, consistently misclassifying all active OOD tasks as `Movie 3` with a high 
% mean confidence of 60.0\%. In %stark 
% contrast, \textbf{\eegtrans{} demonstrated a ``conservative collapse"}, correctly identifying the OOD tasks as unfamiliar by defaulting to its most neutral known class (`Resting State`) with an appropriately low confidence of 26.0\%. This behavior is not a %model 
% failure but a sign of %robust 
% uncertainty quantification, a critical property for safe real-world deployment. Both models correctly classified the in-distribution `Resting State` control task with high accuracy.

\begin{table}[h]
\centering
\caption{Zero-shot cross-task generalization on unseen active tasks (N=245) and an in-distribution control task (N=525).}
\label{tab:cross_task}
\resizebox{\linewidth}{!}{%
\begin{tabular}{llcc}
\toprule
\textbf{Model} & \textbf{Task Input} & \textbf{Dominant Prediction} & \textbf{Mean Confidence} \\
\midrule
\multicolumn{4}{l}{\textit{Out-of-Distribution (Active Tasks)}} \\
\textbf{S5} & Symbol Search & Movie 3 & 60.0\% \\
(Specialized) & Contrast Change & Movie 3 & 60.0\% \\
& Spatial Memory & Movie 3 & 60.5\% \\
\cmidrule{2-4}
\textbf{\eegtrans{}} & Symbol Search & Resting State & 26.0\% \\
(Robust) & Contrast Change & Resting State & 26.0\% \\
& Spatial Memory & Resting State & 26.0\% \\
\midrule
\multicolumn{4}{l}{\textit{In-Distribution Control}} \\
S5 & Resting State & Resting State & $>95.0\%$ \\
\eegtrans{} & Resting State & Resting State & $>90.0\%$ \\
\bottomrule
\end{tabular}
}
\end{table}

% \begin{table}[h]
% \centering
% \caption{Calibration and uncertainty metrics for 64s models on the in-distribution test set. Lower values indicate better calibration.}
% \label{tab:calibration}
% \resizebox{\columnwidth}{!}{%
% \begin{tabular}{lccc}
% \toprule
% \textbf{Model} & \textbf{NLL $\downarrow$} & \textbf{Brier Score $\downarrow$} & \textbf{ECE (\%) $\downarrow$} \\
% \midrule
% S5 & \textbf{0.056 $\pm$ 0.008} & \textbf{0.006 $\pm$ 0.002} & \textbf{1.09 $\pm$ 0.12} \\
% \eegtrans{} & 0.999 $\pm$ 0.023 & 0.077 $\pm$ 0.005 & 13.41 $\pm$ 1.68 \\
% CNN & 0.079 $\pm$ 0.014 & 0.007 $\pm$ 0.002 & 0.96 $\pm$ 0.20 \\
% \bottomrule
% \end{tabular}
% }
% \end{table}

\begin{table}[h]
\centering
\caption{Calibration metrics (64 s, in-distribution). Lower is better.}
\label{tab:calibration}
\begin{tabular}{lccc}
\toprule
Model & NLL $\downarrow$ & Brier $\downarrow$ & ECE (\%) $\downarrow$ \\
\midrule
S5    & $\textbf{0.056} \pm 0.008$ & $\textbf{0.006} \pm 0.002$ & $1.09 \pm 0.12$ \\
EEGXF & $0.999 \pm 0.023$ & $0.077 \pm 0.005$ & $13.41 \pm 1.68$ \\
CNN   & $0.079 \pm 0.014$ & $0.007 \pm 0.002$ & $\textbf{0.96} \pm 0.20$ \\
\bottomrule
\end{tabular}
\end{table}

\section{Discussion and Conclusion}
\label{sec:conclusion}

Our systematic benchmark reveals a critical accuracy-efficiency-robustness trade-off in EEG decoding. In subject-mixed settings, both CNN and S5 architectures demonstrate powerful scaling with temporal context. However, S5 achieves this high performance with over 20x fewer parameters and greater training efficiency. Furthermore, in more challenging subject-independent tests (LOSO), S5's superiority in learning generalizable features becomes even more evident.

However, this efficiency comes at the cost of robustness. Our novel cross-task generalization test demonstrates that S5 is prone to overconfident, incorrect predictions on unseen tasks. Conversely, \eegtrans's ``conservative collapse'' to a neutral, low-confidence state represents superior uncertainty handling. This trade-off forces a practical choice: S5 is ideal for speed in controlled settings, while \eegtrans{} is preferable for robustness.

We acknowledge that this study relies on the HBN dataset to probe long-context scaling; future work should validate these architectural findings on standard benchmarks like Sleep-EDF. Additionally, while shorter segments used single seeds due to computational constraints, the multi-seed evaluation of key long-context results ensures the primary findings are robust.

\clearpage
\bibliographystyle{IEEEbib}
\bibliography{refs}

@misc{shirazi2024hbn,
      title={{The Healthy Brain Network EEG Datasets (HBN-EEG)}}, 
      author={Seyed Yahya Shirazi and Alexandre Franco and Mauricio Scopel Hoffman and Nathalia Esper and Dung Truong and Arnaud Delorme and Michael Milham and Scott Makeig},
      year={2024},
      eprint={2410.02103},
      archivePrefix={arXiv},
      primaryClass={eess.SP}
}

@article{roy2019deep,
  title={Deep learning-based electroencephalography analysis: a systematic review},
  author={Roy, Yannick and Banville, Hubert and Albuquerque, Isabela and Gramfort, Alexandre and Falk, Tiago H and Faubert, Jocelyn},
  journal={Journal of Neural Engineering},
  volume={16},
  number={5},
  pages={051001},
  year={2019},
  publisher={IOP Publishing}
}

@inproceedings{vaswani2017attention,
  author={Vaswani, Ashish and Shazeer, Noam and Parmar, Niki and Uszkoreit, Jakob and Jones, Llion and Gomez, Aidan N and Kaiser, \L{}ukasz and Polosukhin, Illia},
  booktitle={Advances in Neural Information Processing Systems (NeurIPS)},
  editor={I. Guyon and U. V. Luxburg and S. Bengio and H. Wallach and R. Fergus and S. Vishwanathan and R. Garnett},
  pages={5998--6008},
  publisher={Curran Associates, Inc.},
  title={Attention is all you need},
  volume={30},
  year={2017}
}

@misc{gu2023mamba,
      title={{Mamba: Linear-Time Sequence Modeling with Selective State Spaces}}, 
      author={Albert Gu and Tri Dao},
      year={2023},
      eprint={2312.00752},
      archivePrefix={arXiv},
      primaryClass={cs.LG}
}

@article{song2023eeg,
  author={Song, Yonghao and Jia, Zhengyu and Li, Mingsheng},
  journal={IEEE Transactions on Affective Computing}, 
  title={{EEG-Conformer: A Hybrid Framework for Cross-Subject EEG-Based Emotion Recognition}}, 
  year={2023},
  volume={},
  number={},
  pages={1-12},
  doi={10.1109/TAFFC.2023.3239790}
}

@misc{gu2022efficientlymodelinglongsequences,
      title={Efficiently Modeling Long Sequences with Structured State Spaces}, 
      author={Albert Gu and Karan Goel and Christopher Ré},
      year={2022},
      eprint={2111.00396},
      archivePrefix={arXiv},
      primaryClass={cs.LG},
      url={https://arxiv.org/abs/2111.00396}, 
}

@article{wang2025eegmamba,
  title={EEGMamba: An EEG foundation model with Mamba},
  author={Wang, Jiquan and Zhao, Sha and Luo, Zhiling and Zhou, Yangxuan and Li, Shijian and Pan, Gang},
  journal={Neural Networks},
  year={2025},
  doi={10.1016/j.neunet.2025.107816}
}

@inproceedings{siddhad2024drowzee,
  title={DrowzEE-G-Mamba: Leveraging EEG and State Space Models for Driver Drowsiness Detection},
  author={Siddhad, Gourav and Dey, Sayantan and Roy, Partha Pratim},
  booktitle={International Conference on Pattern Recognition (ICPR)},
  series={Lecture Notes in Computer Science},
  volume={15327},
  pages={281--295},
  year={2024},
  publisher={Springer},
  doi={10.1007/978-3-031-52241-1_18}
}

@article{tran2024eegssm,
  title={EEG-SSM: Leveraging State-Space Model for Dementia Detection},
  author={Tran, Xuan-The and Le, Linh and Lin, Chin-Teng},
  journal={arXiv preprint arXiv:2407.17801},
  year={2024},
  doi={10.48550/arXiv.2407.17801}
}

@article{li2025dgmamba,
  title={A Novel State Space Model with Dynamic Graph Neural Network for EEG Event Detection},
  author={Li, Xinying and Yan, Shengjie and Wu, Yonglin and Dai, Chenyun and Guo, Yao},
  journal={International Journal of Neural Systems},
  volume={35},
  number={03},
  pages={2550008},
  year={2025},
  publisher={World Scientific},
  doi={10.1142/S012906572550008X}
}

@article{schirrmeister2017deep,
  title={Deep learning with convolutional neural networks for EEG decoding and visualization},
  author={Schirrmeister, Robin Tibor and Springenberg, Jost Tobias and Fiederer, Lukas Dominique Josef and Glasstetter, Martin and Eggensperger, Katharina and Tangermann, Michael and Hutter, Frank and Burgard, Wolfram and Ball, Tonio},
  journal={Human brain mapping},
  volume={38},
  number={11},
  pages={5391--5420},
  year={2017},
  publisher={Wiley Online Library}
}

\end{document}